\pdfoutput=1

\documentclass[11pt]{article}

\usepackage[]{ACL}
\usepackage{times}
\usepackage{latexsym}
\usepackage[T1]{fontenc}
\usepackage[utf8]{inputenc}
\usepackage{microtype}
\usepackage{inconsolata}
\usepackage{amsmath}
\usepackage[capitalize,noabbrev]{cleveref}
\usepackage{amssymb}
\usepackage{mathtools}
\usepackage{amsthm}
\usepackage{multirow, multicol}
\usepackage{tabularx,tabulary,booktabs}
\usepackage{color,soul}
\usepackage{adjustbox}
\usepackage{bm}
\usepackage{xspace}
\usepackage{arydshln}
\usepackage{colortbl}
\usepackage{enumitem}

\newcommand{\ourmethod}{\textsc{TESS}\xspace}

\title{TESS: Text-to-Text Self-Conditioned Simplex Diffusion}

\author{
Rabeeh Karimi Mahabadi$^{1,4}$\thanks{\hspace{6pt}Co-first authors.} \qquad
Hamish Ivison$^{3,5*}$\thanks{\hspace{6pt}Work done during employment at AI2.} \qquad
Jaesung Tae$^2$ \vspace{2pt} \\
\textbf{James Henderson}$^4$ \qquad
\textbf{Iz Beltagy}$^3$ \qquad
\textbf{Matthew E. Peters}$^3$$^{\dagger\ddagger}$ \qquad \textbf{Arman Cohan}$^{2,3}$\thanks{\hspace{6pt}Equal advising.} \vspace{8pt} \\
$^1$EPFL\quad  $^2$Yale University\quad  $^3$Allen Institute for AI \\  $^4$Idiap Research Institute \quad
$^5$University of Washington \\
\small{\texttt{rabeeh.karimimahabadi@epfl.ch}}, 
\small{\texttt{hamishi@allenai.org}} 
}

\begin{document}
\maketitle

\begin{abstract}
Diffusion models have emerged as a powerful paradigm for generation, obtaining strong performance in various continuous domains. 
However, applying continuous diffusion models to natural language remains challenging due to its discrete nature and the need for a large number of diffusion steps to generate text, making diffusion-based generation expensive.
In this work, we propose \underline{Te}xt-to-text \underline{S}elf-conditioned \underline{S}implex Diffusion (\ourmethod), a text diffusion model that is fully non-autoregressive, employs a new form of self-conditioning, and 
applies the diffusion process on the logit simplex space rather than the learned embedding space.
Through extensive experiments on natural language understanding and generation tasks including summarization, text simplification, paraphrase generation, and question generation, we demonstrate that \ourmethod outperforms state-of-the-art non-autoregressive models, requires fewer diffusion steps with minimal drop in performance, and is competitive with pretrained autoregressive sequence-to-sequence models. We publicly release our codebase.\footnote{\url{https://github.com/allenai/tess-diffusion}}
\end{abstract}

\section{Introduction}

Diffusion models~\citep{sohl2015deep, ho2020denoising, song2021scorebased} have achieved state-of-the-art performance in various continuous domains, such as image~\citep{nichol2021improved}, audio~\citep{kong2020diffwave,Shen2023NaturalSpeech2L}, video~\citep{ho2022video}, and text-to-image generation~\citep{saharia2022photorealistic, ramesh2022hierarchical}. 
Inspired by the success of diffusion for continuous domains, recent works have adapted diffusion to discrete spaces, such as text \cite{austin2021structured,hoogeboom2021argmax,savinov2021step,reid2022diffuser}.
One line of work proposes diffusing the model latent space by adding Gaussian noise to input word embeddings \cite{li2022diffusion}. Another approach, SSD-LM~\citep{han2022ssd}, adds noise to the vocabulary probability simplex. 

Direct diffusion on the probability simplex is desirable \cite{richemond2022categorical} as it eliminates the need for an extra step to map diffused embeddings to actual discrete inputs or auxiliary methods such as binary encoding \cite{chen2022analog}. Despite its strong performance, however, SSD-LM has several shortcomings: a lack of self-conditioning \cite{chen2022analog}, a lack of extensive evaluation on downstream tasks, and most notably, its restriction to generating blocks of 25 tokens, which hinders the potential benefits of full diffusion, e.g., the ability to perform arbitrary infilling, flexible generation, and a global view of the sequence.


In this work, we present \ourmethod, a 
text-to-text diffusion model, which overcomes several limitations of prior works: restrictions on scale~\cite{hoogeboom2021argmax,austin2021structured}, dependence on pretrained embeddings \cite{strudel2022self}, semi-autoregressive nature \cite{han2022ssd}, and short generation length~\cite{gong2022diffuseq}.
\ourmethod closely follows \citet{han2022ssd,Han2023SSD2SA} by performing diffusion on the vocabulary logit space rather than the typical embedding space. Unlike SSD-LM, however, \ourmethod is fully non-autoregressive and performs diffusion on the entire sequence. It also incorporates a novel form of self-conditioning, which demonstrates a competitive edge over the original self-conditioning method \cite{chen2022analog} and dramatically improves the efficiency and quality of TESS.

We evaluate \ourmethod on a suite of natural language generation (NLG) tasks including summarization, text simplification, paraphrase generation, and question generation. Our empirical results surpass the current state-of-the-art non-autoregressive and diffusion-based approaches and are on par with a strong pretrained encoder-decoder language model~\cite{lewis2020bart}. In particular, our simplex-based self-conditioning method substantially improves generation quality. We also evaluate \ourmethod on natural language understanding (NLU) tasks from the GLUE benchmark \cite{wang2018glue} and show that it performs comparably to strong masked language model baselines. Our contributions can be summarized as follows.

\begin{enumerate}[wide=0pt]
    \setlength\itemsep{0em}
    \item We demonstrate the effectiveness of a fully non-autoregressive scheme for text diffusion models, which outperforms strong autoregressive and non-autoregressive baselines.
    \item We propose a new self-conditioning method that exploits the simplex semantics of the diffusion space and greatly improves performance.
    \item We evaluate \ourmethod on a suite of diverse NLG and NLU tasks, highlighting the effectiveness of our text-to-text simplex diffusion paradigm.
    \item We show \ourmethod' fully non-autoregressive approach results in faster and more efficient sampling than semi and fully autoregressive methods for long sequences.
\end{enumerate}

We will release our trained models and code to promote open research in the field of diffusion-based text generation.

\section{Background}

We revisit continuous diffusion models~\citep{sohl2015deep}, following the formulation of Denoising Diffusion Models~\citep{ho2020denoising,song2020denoising}.
\paragraph{Training} Given a sample $\mathbf{x}_0 \in \mathbb{R}^d$ from a data distribution $p_\text{data}$, a forward diffusion process $q(\mathbf{x}_t | \mathbf{x}_{t - 1})$ is a Markov chain that generates a sequence of latent variables $\mathbf{x}_1, \dots, \mathbf{x}_T$ by gradually adding Gaussian noise at each time step $t \in \{1, 2, \dots, T \}$ with variance $\beta_{t} \in \mathbb{R}_{>0}$:
\begin{align}
q(\mathbf{x}_t | \mathbf{x}_{t - 1}) = \mathcal{N}(\mathbf{x}_t; \sqrt{1-\beta_t} \mathbf{x}_{t - 1}, \beta_t \mathbf{I}).
\end{align}
Let $\bm{\epsilon}_t \sim \mathcal{N}(0, \mathbf{I})$, $\alpha_t = 1 - \beta_t$, and $\bar{\alpha}_t = \prod_{s=1}^t \alpha_s$. Then sampling $\mathbf{x}_t$ at an arbitrary time step $t$ has the closed-form solution
\begin{align}
\mathbf{x}_t = \sqrt{\bar{\alpha}_t} \mathbf{x}_0 + \sqrt{1-\bar{\alpha}_t} \bm{\epsilon}_t.
\label{eq:ddpm-forward}
\end{align}
Given a well-behaved noise schedule $\{ \beta_t \}^T_{t = 1}$, $\mathbf{x}_T$ follows a stationary prior distribution $\mathcal{N}(0, \mathbf{I})$. Therefore, if we can approximate the reverse process $q(\mathbf{x}_{t - 1} | \mathbf{x}_t, \mathbf{x}_0)$ via a model $p_{\bm{\theta}}(\mathbf{x}_{t - 1} | \mathbf{x}_t)$ with parameters $\bm{\theta}$, then we can sample random noise from a standard Gaussian and gradually denoise it to sample from $p_\text{data}$. In our settings, our model $p_{\bm{\theta}}$ is a transformer model\footnote{Specifically, we use a RoBERTa model~\citep{roberta}, but our formulation could be applied to any transformer variant.}. The reverse process is thus parametrized as
\begin{align}
p_{\theta}(\mathbf{x}_{t - 1} | \mathbf{x}_t) = \mathcal{N}(\bm{\mu_{\theta}}(\mathbf{x}_t, t), \bm{\Sigma_{\theta}}(\mathbf{x}_t, t)).
\end{align}
The model is trained by minimizing the mean squared error between the ground-truth data $\mathbf{x}_0$ and its estimate $\hat{\mathbf{x}}_{\bm{\theta}}$:\footnote{Alternatively, we can train the model to predict the added noise; see~\citet{ho2020denoising}. See also~\citet{song2021scorebased} for a score-matching interpretation.}
\begin{align}
\mathcal{L} = \mathbb{E}_{t,q(\mathbf{x}_0), q(\mathbf{x}_t | \mathbf{x}_0)} \| \mathbf{x}_0 - \hat{\mathbf{x}}_{\bm{\theta}}(\mathbf{x}_t, t)\|^2.
\label{eq:l2_loss}
\end{align}

\paragraph{Noise schedule} The forward diffusion process is defined by a noise schedule. In this work, we follow the cosine schedule~\citep{nichol2021improved} for $\alpha_t$: 
\begin{align}
\bar{\alpha}_t = \frac{f(t)}{f(0)}, \quad f(t) = \cos\left(\frac{t/T + s}{1+s}.\frac{\pi}{2} \right)^2.
\end{align}

\paragraph{Inference} In~\citet{song2020denoising}, model predictions are iteratively denoised for $t = T,\dots, 1$ starting from pure noise, following
$$
\mathbf{x}_{t - 1} = \sqrt{\alpha_{t - 1}} \hat{\mathbf{x}}_{\bm{\theta}} + \sqrt{1 - \alpha_{t - 1}} \cdot \frac{\mathbf{x}_t - \sqrt{\alpha_t} \hat{\mathbf{x}}_{\bm{\theta}}}{\sqrt{1 - \alpha_t}}.
$$
We follow the recently proposed simplex-based diffusion procedure by \citet{han2022ssd}, which allows us to apply diffusion to text without employing auxiliary methods that map categorical data to continuous space \cite{richemond2022categorical}.

\begin{figure*}[t]
\centering
\includegraphics[scale=0.5]{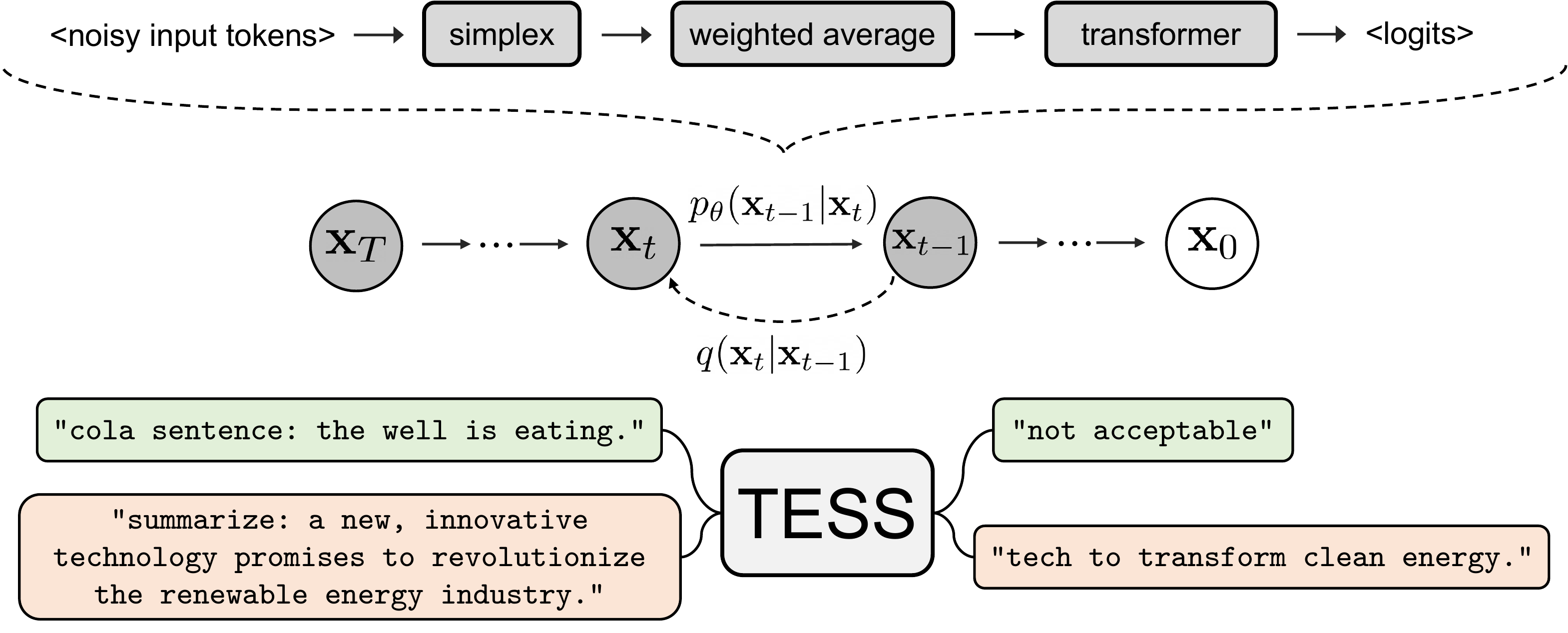}
\caption{Overview of \ourmethod. During training (top), we first add noise to the vocabulary probability simplex, compute a weighted average word embedding, and denoise it using a transformer encoder. To generate from our model, we begin with noise and iteratively refine it into a final logit distribution (middle). The resulting model can be used for a wide range of NLG and NLU end tasks (bottom).}
\label{fig:our_method}
\end{figure*} 

\section{Method} \label{sec:method}
In this section, we present \ourmethod, a simplex diffusion-based text-to-text model. Building upon SSD-LM \cite{han2022ssd}, we propose a fully non-autoregressive model with self-conditioning. 

\paragraph{Continuous data representation} Let $\mathcal{V}$ denote the vocabulary space. Following~\citet{han2022ssd}, we map the ID of each token to be generated $w \in \mathcal{V}$ to a $k$-logit simplex to produce $\mathbf{s}^w \in \{\pm k\}^{|\mathcal{V}|}$, whose $i$-th component satisfies
\begin{equation}
s^w_{(i)} = 
\begin{cases}
k, & \text{if}\quad i = w, \\
-k, & \text{otherwise},
\label{eq:scaled-simplex}
\end{cases}
\end{equation}
with a hyperparameter $k \in \mathbb{R}$. We then produce a probability simplex over $\mathcal{V}$ via $\mathbf{p}^w = \text{softmax}(\mathbf{s}^w)$. Finally, we compute the weighted sum of word embeddings to obtain a continuous embedding vector, $\mathbf{h}^w = \mathbf{E} \mathbf{p}^w$, where $\mathbf{E} \in \mathbb{R}^{d \times |\mathcal{V}|}$ is the word embedding matrix, $d$ denotes the size of the hidden dimension, and $\mathbf{h}^w \in \mathbb{R}^d$.   %

\paragraph{Time step embeddings}
After computing the continuous word embeddings, we add the time step embeddings to inform the model of the current time step. Our time step embedding is a linear layer, and we feed scaled time steps $t / T$ to this layer. The output is a time step embedding in $\mathbb{R}^d$ that is added to $\mathbf{h}_w$ to produce the final latent input vector.

\paragraph{Text-to-text non-autoregressive modeling}
Unlike SSD-LM, which feeds small blocks of text to semi-autoregressively generate sequences of text, we feed the entire latent vector along with the context into an encoder transformer model. This is a key difference between our approach and SSD-LM, as it allows for a fully non-autoregressive model capable of generating sequences of any length. In practice, our evaluation tasks often require output sequences of 100 tokens or more, and by moving to a fully non-autoregressive paradigm, we are able to generate entire output sequences in parallel without resorting to semi-autoregressive generation.  

\paragraph{Forward diffusion}
Let $\mathbf{w} = (w_1, \dots, w_L)$ be a sentence of $L$ tokens such that $w_i \in \mathcal{V}$, and $\mathbf{S}_0 = (\mathbf{s}^{w_1}, \dots, \mathbf{s}^{w_L}) \in \{\pm k\}^{L \times |\mathcal{V}|}$ be the $k$-logit simplex representation of $\mathbf{w}$. %
We add noise to the $k$-logit simplex representation during training according to
\begin{align}
\mathbf{S}_t = \sqrt{\bar{\alpha}_t} \mathbf{S}_0 + \sqrt{1-\bar{\alpha}_t} \bm{\epsilon}_t, 
\label{eq:forward_diff}
\end{align}
where subscript denotes the time step and $\bm{\epsilon}_t \sim \mathcal{N}(0, k^2 \mathbf{I})$.

\paragraph{Training} Typical diffusion models are trained with mean squared error loss as in Equation \eqref{eq:l2_loss} to predict the ground-truth data. This objective is known to be unstable for text diffusion models~\citep{dieleman2022continuous}. \citet{strudel2022self} froze word embeddings and used specific scaling to deal with training instability. In this work, following~\citet{han2022ssd}, we instead compute the usual cross-entropy loss between the ground-truth tokens $\mathbf{w}$ and the model prediction given a noisy logit simplex $\mathbf{S}_t$ at time step $t$.
\begin{align}
\mathcal{L}
&= \mathbb{E}_{t,q(\mathbf{S}_0), q(\mathbf{S}_t | \mathbf{S}_0)} \left[ -\sum_{i = 1}^{L} \log p_{\bm{\theta}}(w_i | \mathbf{S}_t, t) \right].
\end{align}

\paragraph{Sampling} During inference, we sample $\mathbf{S}_T$ from the prior $\mathcal{N}(0, k^2 \mathbf{I})$ and run the reverse process for $t = T,\dots, 1$ on the noisy $k$-logit simplex.
The reverse process can be approximated via
\begin{align}
\mathbf{S}_{t - 1} = \sqrt{\bar{\alpha}_{t-1}} \hat{\mathbf{S}}_{\bm{\theta}}(\mathbf{S}_t, t) + \sqrt{1-\bar{\alpha}_{t-1}} \bm{\epsilon}_t. \label{eq:rev_diff}
\end{align}
See Appendix~\ref{app:inference-step} for details. This resembles the forward process in Equation~\eqref{eq:forward_diff}, which allows for an intuitive interpretation: to reverse one step from $t$, we take the model prediction $\hat{\mathbf{S}}_{\bm{\theta}}$ as the hypothetical ground-truth, then corrupt it by $(t - 1)$ time steps. To construct the model prediction, we project the logits predicted by the underlying encoder model via argmax as a pseudo-inverse of Equation~\eqref{eq:scaled-simplex} to match the initial $k$-logit representation:
\begin{equation}
\hat{s}^{w}_{(i)} = 
\begin{cases} 
k, & \text{if\quad} i= \text{argmax}(\mathbf{s}^{w}), \\
-k, & \text{otherwise}.
\label{eq:scaled-simplex2}
\end{cases}
\end{equation}

\paragraph{Self-conditioning} 
In typical diffusion models, the model predicts the original data $\mathbf{x}_0$ conditioned on its corrupted version, i.e., $\hat{\mathbf{x}}_0^t = \hat{\mathbf{x}}_{\bm{\theta}}(\mathbf{x}_t, t)$, where $\hat{\mathbf{x}}_0^t$ denotes the estimate of $\mathbf{x}_0$ at time step $t$. In this setting, the model's estimates at previous time steps are discarded. However, in self-conditioning~\citep{chen2022analog}, the model conditions its prediction on both $\mathbf{x}_t$ and its previously generated output, i.e., $\hat{\mathbf{x}}_0^t = \hat{\mathbf{x}}_{\bm{\theta}}(\mathbf{x}_t, \hat{\mathbf{x}}_0^{t + 1}, t)$. To adapt the model for self-conditioning, we stochastically zero out the self-condition such that
\begin{equation}
\hat{\mathbf{x}}_0^t = 
\begin{cases}
\hat{\mathbf{x}}_{\bm{\theta}}(\mathbf{x}_t, \hat{\mathbf{x}}_0^{t + 1}, t),& \text{with probability $\rho$}\\
\hat{\mathbf{x}}_{\bm{\theta}}(\mathbf{x}_t, 0,
t), & \text{otherwise},
\label{eq:scaled-simplex3}
\end{cases}
\end{equation}
where the self-conditioning previous prediction is computed as $\hat{\mathbf{x}}_0^{t + 1} = \hat{\mathbf{x}}_{\bm{\theta}}(\mathbf{x}_{t + 1}, 0, t + 1)$, with gradients detached. 
We set $\rho = 0.5$ during training; during inference, we always use self-conditioning ($\rho = 1$). 

We propose a new self-conditioning method that exploits the simplex nature of our diffusion space. Let $\mathbf{s}_t \in \mathbb{R}^{|\mathcal{V}|}$ be a noised $k$-logit simplex for an arbitrary token $w$.\footnote{We write $\mathbf{s}^w_t$ as $\mathbf{s}_t$ for brevity.} Instead of concatenating the previous prediction with $\mathbf{s}_t$ and re-projecting, we first compute the average of simplex probabilities 
\begin{align}
\mathbf{p}^w_\text{avg} = \frac12 \left(\text{softmax}(\mathbf{s}_t) + \text{softmax}(\hat{\mathbf{s}}^{t + 1}_0) \right).
\end{align}
Note that $\mathbf{p}^w_\text{avg}$ is a well-defined categorical distribution over $\mathcal{V}$. We then compute a continuous embedding vector, $\mathbf{h}^w = \mathbf{E} \mathbf{p}^w_\text{avg}$, and use this vector as input to our underlying model to make a prediction for the given diffusion step following Equation~\ref{eq:rev_diff}.
This is more efficient than the original self-conditioning method, which projects down the concatenated vectors. In Section \textsection\ref{sec:ablations}, we also demonstrate the empirical effectiveness of this method over the original.

\paragraph{Variable sequence length}
A notable challenge in non-autoregressive generation is the assumption of fixed sequence lengths during inference. To overcome this issue, we follow prior work in embedding-space diffusion by using padding tokens \cite{li2022diffusion}.
Specifically, during training, we always pad the variable-length output sequence to a fixed length using padding tokens. These padding tokens are included when computing the cross-entropy loss so that \ourmethod learns to generate them. During inference, we specify the maximum sequence length and run sampling as usual.

\section{Experiments} \label{sec:experiments}


\subsection{Tasks and Datasets} 

\paragraph{Paraphrase generation} This task involves rephrasing a sentence while maintaining the semantics of the original.
We use Quota Question Pairs (QQP),\footnote{\url{https://www.kaggle.com/c/quora-question-pairs}} which is composed of 147K positive pairs. We use only the positively-labelled pairs, which have the same meaning.

\paragraph{Text simplification} This task involves simplifying complex sentences while retaining their original meaning. 
We use the NEWSELA-AUTO dataset~\citep{jiang2020neural}, which is composed of 666K complex-simplified sentences. %

\paragraph{Question generation} This task involves generating a question given an input context.
We use the QUASAR-T dataset~\citep{dhingra2017quasar} processed by~\citet{yuan2022seqdiffuseq}, resulting in 119K document-question pairs. 

\paragraph{Summarization} 
We evaluate our method on the CNN-DailyMail dataset~\citep{hermann2015teaching}, which comprises 300K articles and summaries.

\paragraph{Classification} We consider a set of classification tasks in the GLUE benchmark~\citep{wang2018glue} covering a variety of tasks, including paraphrase detection (MRPC, QQP), sentiment classification (SST-2), natural language
inference (MNLI,\footnote{We report the accuracy on the matched validation set.} RTE, QNLI), and linguistic acceptability (CoLA).\footnote{Following~\citet{devlin-etal-2019-bert, raffel2019exploring}, as a
common practice and due to the adversarial nature of WNLI, we do not experiment with WNLI.}

\subsection{Baselines}
We compare \ourmethod to several autoregressive baselines as well as state-of-the-art text diffusion models. For autoregressive methods, we consider GPT-2~\cite{radford2019language}, BART~\cite{lewis2020bart}, and GPVAE-T5~\cite{du2022diverse}, a latent-structured variable model and an extension to T5 \cite{raffel2019exploring}. For text diffusion models, we consider Diffuser \cite{reid2022diffuser}, DiffuSeq \cite{gong2022diffuseq}, SeqDiffuSeq \cite{yuan2022seqdiffuseq}, SUNDAE~\citep{savinov2021step}, LevT~\citep{gu2019levenshtein}, a widely used iterative non-autoregressive model, and SSD-LM~\citep{han2022ssd} initialized from the same pretrained RoBERTa model as \ourmethod and trained using the official SSD-LM codebase.\footnote{\url{https://github.com/xhan77/ssd-lm}} We report results without using additional decoding methods such as minimum Bayes risk decoding. We provide further details on baseline results in Appendix~\ref{app:experimental_detials}.

\begin{table}[t]
\centering
\footnotesize
\setlength\tabcolsep{0pt}
\begin{adjustbox}{max width=\linewidth}
\begin{tabular*}{1.1\linewidth}{@{\extracolsep{\fill}}lcccc@{}}
\toprule
\multirow{1}{*}{\textbf{Model}} & \multicolumn{4}{c}{{\textbf{Paraphrase Generation}}}\\
\cmidrule(lr){2-5}
 & \textbf{BLEU} & \textbf{BERT} & \textbf{R-L} & \textbf{D-1/4} \\
\midrule
\multicolumn{5}{@{}l}{\it \textbf{Autoregressive Models}} \\
[0.5ex]\hdashline[1pt/1pt]\noalign{\vskip 0.5ex}
BART \cite{lewis2020bart} & \underline{30.4}& \underline{85.7} &61.4 & \underline{98.8}/61.0\\[0.5ex]\hdashline[1pt/1pt]\noalign{\vskip 0.5ex}
GPT-2\textsubscript{base}$^\dagger$ \cite{radford2019language} & 19.8  & 82.5 & 52.1 & 98.0/\underline{62.5}\\
GPT-2\textsubscript{large}$^\dagger$ \cite{radford2019language}   &20.6 & 83.6 & 54.2 & 98.2/50.2\\
GPVAE-T5$^\dagger$ \cite{du2022diverse}  & 24.1 & 84.7 & 58.9 & 96.9/61.7  \\
\midrule 
\multicolumn{5}{@{}l}{\it \textbf{Non-Autoregressive Models}}\\
\midrule 
LevT$^\dagger$ \cite{gu2019levenshtein}  & 22.7 & 83.4 &57.9 & 97.9/33.3 \\
\midrule
\multicolumn{5}{@{}l}{\it \textbf{Non-Autoregressive Diffusion Models}}\\
[0.5ex]\hdashline[1pt/1pt]\noalign{\vskip 0.5ex}
DiffuSeq$^\star$~\citep{gong2022diffuseq} & 18.5 & 79.5 & --- & 97.6/---\\
SeqDiffuSeq$^\star$~\citep{yuan2022seqdiffuseq} & 23.3 & 82.9 & --- & 98.1/---\\
SSD-LM~\citep{han2022ssd} & 22.9 & 83.8 & 58.3 & \underline{\bf{98.8}}/57.3 \\
\midrule 
\ourmethod (Ours) & \bf{30.2} & \underline{\bf{85.7}} & \underline{\bf{62.2}} & 98.5/\bf{61.1} \\  %
\bottomrule
\end{tabular*}
\end{adjustbox}
\caption{Results on Paraphrase Generation task. $^\dagger$ indicates results from from~\citet{gong2022diffuseq}, * indicates results from~\citet{yuan2022seqdiffuseq}. Boldfaced results show the best across all non-AR models; underlined results are the best across all models.}
\label{tbl:paraphrase}
\end{table}

\subsection{Evaluation}
For summarization, we report ROUGE-1 (R1), ROUGE-2 (R2), and ROUGE-L (R-L) variants~\cite{rouge} as done in prior text summarization work \cite{lewis2020bart}. We quantify both generation quality and diversity. For evaluating generation quality, we report BLEU \cite{bleu}, ROUGE-L \cite{rouge} and BERTScore \cite{zhang2019bertscore} following \citet{gong2022diffuseq} and \citet{yuan2022seqdiffuseq}. For evaluating diversity, we report distant unigrams (D-1) and diverse 4-grams (D-4) \cite{Deshpande2018FastDA}. For text simplification, we use the standard SARI \cite{xu2016optimizing}, and following \citet{gong2022diffuseq, yuan2022seqdiffuseq}, we also include BLEU, BERTScore, and ROUGE-L. 

\subsection{Implementation}
We start from the RoBERTa pretrained checkpoint
\citep{roberta} and finetune the model on downstream tasks using our proposed self-conditioned simplex diffusion method. The number of diffusion sampling steps at inference time is set to $T=1000$ for generation and $T=10$ for classification tasks. During training, we use $T=5000$. We set the simplex scale to $k=5$. Additional details are listed in Appendix~\ref{app:experimental_detials}.

\begin{table}[t]
\centering
\footnotesize
\setlength\tabcolsep{0pt}
\begin{adjustbox}{max width=\linewidth}
\begin{tabular*}{1.1\linewidth}{@{\extracolsep{\fill}}lcccc@{}}
\toprule
\multirow{1}{*}{\textbf{Model}} & \multicolumn{4}{c}{{\textbf{Text Simplification}}}\\
\cmidrule(lr){2-5}
& \textbf{SARI} & \textbf{BLEU} & \textbf{BERT} & \textbf{R-L} \\
\midrule
\multicolumn{5}{@{}l}{\it \textbf{Autoregressive Models}} \\
[0.5ex]\hdashline[1pt/1pt]\noalign{\vskip 0.5ex}
BART \cite{lewis2020bart} & 49.9&41.4&81.7 & 58.1 \\
GPT-2\textsubscript{base}$^\dagger$ \cite{radford2019language}   & ---& 30.8 & 80.2 & 54.6\\
GPT-2\textsubscript{large}$^\dagger$ \cite{radford2019language} & ---& 26.9 & 78.8 &  51.1\\
GPVAE-T5$^\dagger$\cite{du2022diverse} &  --- & 33.9 & 81.7 & 58.3 \\
\midrule 
\multicolumn{5}{@{}l}{\it \textbf{Non-Autoregressive Models}}\\
\midrule 
LevT$^\dagger$ \cite{gu2019levenshtein}  & ---& 20.5 & 72.5 & 44.0\\
\midrule
\multicolumn{5}{@{}l}{\it \textbf{Non-Autoregressive Diffusion Models}}\\
[0.5ex]\hdashline[1pt/1pt]\noalign{\vskip 0.5ex}
DiffuSeq$^\star$\footnotesize{~\citep{gong2022diffuseq}} & ---  & 29.9  & 79.1 & --- \\
SeqDiffuSeq$^\star$~\citep{yuan2022seqdiffuseq} & --- & 37.1 & 82.1 & --- \\
SSD-LM~\citep{han2022ssd} & 36.3 & 12.5 & 69.5 & 39.6 \\
\midrule 
\ourmethod (Ours) &   \underline{\bf{54.3}}& \underline{\bf{41.5}} & \underline{\bf{82.1}}  & \underline{\bf{59.4}}  \\ %
\bottomrule
\end{tabular*}
\end{adjustbox}
\caption{Results on the text simplification task.  $^\dagger$ indicates results from from~\citet{gong2022diffuseq}, * indicates results from~\citet{yuan2022seqdiffuseq}.
}
\label{tbl:simplification-base}
\end{table}

\section{Results}
\label{sec:results}
\subsection{Paraphrase Generation}
As seen in Table~\ref{tbl:paraphrase}, \ourmethod significantly outperforms GPT-2 and other non-autoregressive and diffusion baselines in quality metrics (BLEU, BERT, and ROUGE) while achieving parity in diversity metrics (D-1/D-4). Moreover, \ourmethod obtains competitive overall performance with BART. Note that BART uses a denoising pretraining objective, which is substantially conducive to sequence-to-sequence tasks~\citep{lewis2020bart}; we do not perform any additional pretraining beyond RoBERTa's checkpoint, which was only pretrained on the general masked language modeling objective. We suspect that \ourmethod could significantly benefit from additional diffusion pretraining (see Section \S\ref{sec:future}).

\subsection{Text Simplification}
\label{subsec:simplification}
Results of the text simplification task on the NEWSELA-AUTO dataset are presented in Table~\ref{tbl:simplification-base}. %
\ourmethod outperforms all baselines typically by large margins, including both autoregressive and non-autoregressive models.

\subsection{Question Generation}
As shown in Table~\ref{tbl:question-generation}, \ourmethod outperforms other diffusion and non-autoregressive models in terms of both quality of generation (BLEU, BERTScore, ROUGE) and diversity (D-1/D-4). It also consistently outperforms other autoregressive baselines except for BART, whose performance is closely matched by \ourmethod. We also train and evaluate \ourmethod without initializing from pretrained RoBERTa (random init), and find that this outperforms all NAR baselines in BLEU and D-1, while remaining close in performance in BERTScore and ROUGE-L. This shows that the TESS framework outperforms baselines even without the benefit of making use of existing pretrained models.

\begin{table}[t]
\centering
\footnotesize
\setlength\tabcolsep{0pt}
\begin{adjustbox}{max width=\linewidth}
\begin{tabular*}{1.1\linewidth}{@{\extracolsep{\fill}}lcccc@{}}
\toprule
\multirow{1}{*}{\textbf{Model}} & \multicolumn{4}{c}{{\textbf{Question Generation}}}\\
\cmidrule(lr){2-5}
 & \textbf{BLEU} & \textbf{BERT} & \textbf{R-L} & \textbf{D-1/4} \\
\midrule
\multicolumn{5}{@{}l}{\it \textbf{Autoregressive Models}} \\
[0.5ex]\hdashline[1pt/1pt]\noalign{\vskip 0.5ex}
BART \cite{lewis2020bart} & 17.4 & \underline{66.2} & 38.8 & \underline{98.2}/61.7  \\
GPT-2\textsubscript{base}$^\dagger$ \cite{radford2019language} & 7.4 & 60.5 & 27.2 & 96.0/\underline{92.2}\\
GPT-2\textsubscript{large}$^\dagger$ \cite{radford2019language} &11.1 & 63.5 & 32.2 & 96.7/80.6 \\
GPVAE-T5$^\dagger$\cite{du2022diverse} &12.5 & 63.1 & 33.9 & 93.8/72.8 \\
\midrule 
\multicolumn{5}{@{}l}{\it \textbf{Non-Autoregressive Models}}\\
\midrule 
LevT$^\dagger$\cite{gu2019levenshtein} &9.3 & 54.9 & 28.9 & 89.1/47.8 \\
\midrule
\multicolumn{5}{@{}l}{\it \textbf{Non-Autoregressive Diffusion Models}}\\
[0.5ex]\hdashline[1pt/1pt]\noalign{\vskip 0.5ex}
DiffuSeq$^\star$~\citep{gong2022diffuseq} & 15.8 & 59.4 & --- & 91.1/---\\
SeqDiffuSeq$^\star$~\citep{yuan2022seqdiffuseq} & 17.2 & 61.4 & ---  & 92.7/---\\
SSD-LM~\citep{han2022ssd} & 14.1 & 62.8 & 38.5 & 94.5/56.9 \\
\midrule 
\ourmethod (random init) & 19.0 & 60.8 & 36.1 & 96.1/62.4 \\ 
\ourmethod (Ours) & \underline{\bf{19.5}} & \bf{65.8} & \underline{\bf{38.9}} & \bf{97.1}/\bf{63.0} \\ 
\bottomrule
\end{tabular*}
\end{adjustbox}
\caption{Results on Question Generation task. 
}
\label{tbl:question-generation}
\end{table} %

\subsection{Summarization}
As shown in Table~\ref{tbl:summarization}, \ourmethod achieves competitive results with BART while outperforming prior diffusion work, Diffuser~\citep{reid2022diffuser}, by 1.9 ROUGE-L points, and its bootstrapped variants by 0.8 ROUGE-L points. Note that additional bootstrapping is orthogonal to their method and can be applied to \ourmethod as well. Additionally, \ourmethod outperforms GENIE, another prior diffusion-based method, while using half the number of diffusion steps. This suggests \ourmethod' simplex-based formulation leads to better performance than alternate diffusion approaches.

\begin{table}[t]
\centering
\footnotesize
\begin{tabular}{@{}lrrr@{}}
\toprule
\multirow{2}{*}{\textbf{Model}} %
&\multicolumn{3}{c}{{\textbf{CNN-DM}}}\\
\cmidrule(lr){2-4} 
 & \textbf{R1} & \textbf{R2} & \textbf{R-L} \\
\midrule
\multicolumn{4}{@{}l@{}}{\it \textbf{Autoregressive Models}} \\
[0.5ex]\hdashline[1pt/1pt]\noalign{\vskip 0.5ex} 
BART \cite{lewis2020bart} & \underline{42.9}& \underline{20.1}&\underline{40.1}\\ %
Transformer~\citep{vaswani2017attention}$^\diamond$ & --- & --- & 36.8\\
\midrule 
\multicolumn{4}{@{}l}{\it \textbf{Non-Autoregressive Diffusion Models}}\\
[0.5ex]\hdashline[1pt/1pt]
SUNDAE~\citep{savinov2021step}$^\diamond$ & ---& --- &  37.0 \\
Diffuser~\citep{reid2022diffuser}$^\diamond$ & --- & --- & 37.8 \\
Diffuser+ AR bootstrap$^\diamond$ &--- & --- & 38.4 \\
Diffuser + source bootstrap$^\diamond$ & ---& --- & 38.9\\
GENIE~\citep{genie} & 41.8 & 18.3 & 35.5 \\
\midrule 
\ourmethod (Ours) &\bf{42.3}&\bf{19.4}&\bf{39.7}\\ %
\bottomrule
\end{tabular}
\caption{Results on CNN-DailyMail dataset. Baseline values marked with $^\diamond$ are taken from~\citet{reid2022diffuser}.
}
\label{tbl:summarization}
\end{table}

\subsection{Text Classification} \label{sec:glue}

To our knowledge, \ourmethod is the first model that is evaluated on both NLG and NLU. We evaluate \ourmethod on classification tasks and directly compare our diffusion-based finetuning method with standard finetuning methods for supervised learning. To perform a controlled experiment, we compare \ourmethod with a similar-sized RoBERTa, which we use to initialize our model. Note that since \ourmethod is text-to-text, similar to T5~\citep{raffel2019exploring}, it can naturally handle classification tasks by generating class labels without the need for verbalizers. For STS-B, which is a regression problem, we recast it as a 21-class classification problem following~\citet{raffel2019exploring}. Results are shown in Table \ref{tab:glue_results_original_data}. We observe that \ourmethod matches or outperforms finetuned RoBERTa on several tasks, achieving roughly 2-point gains on MRPC and RTE.  %

\begin{table*}[t!]
\centering 
\footnotesize
\renewcommand{\arraystretch}{1.2}
\begin{tabular}{@{}l@{\hskip 0.05in}l@{\hskip 0.1in}l@{\hskip 0.1in}l@{\hskip 0.1in}l@{\hskip 0.1in}l@{\hskip 0.1in}l@{\hskip 0.1in}l@{\hskip 0.1in}l@{\hskip 0.1in}cr@{}}
\toprule 
\textbf{Method} & \textbf{MNLI} & \textbf{QNLI} & \textbf{QQP} & \textbf{RTE} & \textbf{SST-2} & \textbf{MRPC} & \textbf{CoLA} & \textbf{STS-B} & \textbf{WNLI} & \textbf{Average} \\
\midrule 
RoBERTa\textsubscript{large} & \bf{90.2}/\bf{90.2} & \bf{94.7} & \bf{92.2} & 86.6 & \bf{96.4} & 90.9 & \bf{68.0} & \bf{92.4} & \bf{91.3} & \bf{88.9} \\ %
TESS\textsubscript{large} (Ours) & 90.1/89.8 & 94.2 & 89.1 & \bf{88.5} & \bf{96.4} & \bf{93.1} & 67.7 & 88.9 & 83.1 & 88.5 \\ %
\bottomrule
\end{tabular}
\caption{Comparison of \ourmethod and RoBERTa on GLUE tasks on the development set.
Following~\citet{devlin-etal-2019-bert}, for  MRPC and QQP, we report F1 score; STS-B, Spearman correlation coefficient; CoLA, Matthews correlation. For all other tasks, we report accuracy. Bold fonts indicate the best results.
}
\label{tab:glue_results_original_data} %
\end{table*} 

\section{Analysis}

\begin{table}[t] 
\centering
\footnotesize
\setlength\tabcolsep{0pt}
\begin{tabular*}{\linewidth}{@{\extracolsep{\fill}}lcccc@{}}
\toprule
\multirow{1}{*}{\textbf{Model}} &  \multicolumn{4}{c}{{\textbf{Text Simplification}}}\\
\cmidrule(lr){2-5} %
 & \textbf{SARI} & \textbf{BLEU} & \textbf{BERT} & \textbf{R-L} \\
\midrule
\ourmethod & 44.1 & 30.8 & 78.8 & 52.6 \\ %
\quad+{orig. self-cond} & 52.2 & \bf{43.3} & 81.9 & 58.9  \\
\quad+{proposed self-cond} & \bf{54.2} & 40.8 &  \bf{82.0} & \bf{59.3}  \\
\midrule %
& \multicolumn{4}{c}{{\textbf{Paraphrase generation}}}\\
 \cmidrule(lr){2-5}
 & \textbf{BLEU} & \textbf{BERT} & \textbf{R-L} & \textbf{D-1/D-4} \\
\midrule 
\ourmethod & 25.9&84.4&59.7&\textbf{98.7}/60.4 \\ %
\quad+{orig. self-cond} & 28.4&85.5&\bf{61.7}&98.6/60.6  \\ %
\quad+{proposed self-cond} & \textbf{29.2}&\bf{85.5}&61.2&98.5/\textbf{61.4}   \\ %
\bottomrule
\end{tabular*}
\caption{Ablation on the effects of self-conditioning. We compare our proposed self-conditioning to the original method in~\citet{chen2022analog}, and the model without self-conditioning. Bold fonts indicate the best results.}
\label{tab:self} %
\end{table}

\begin{figure}[t]
\centering
\includegraphics[scale=0.4]{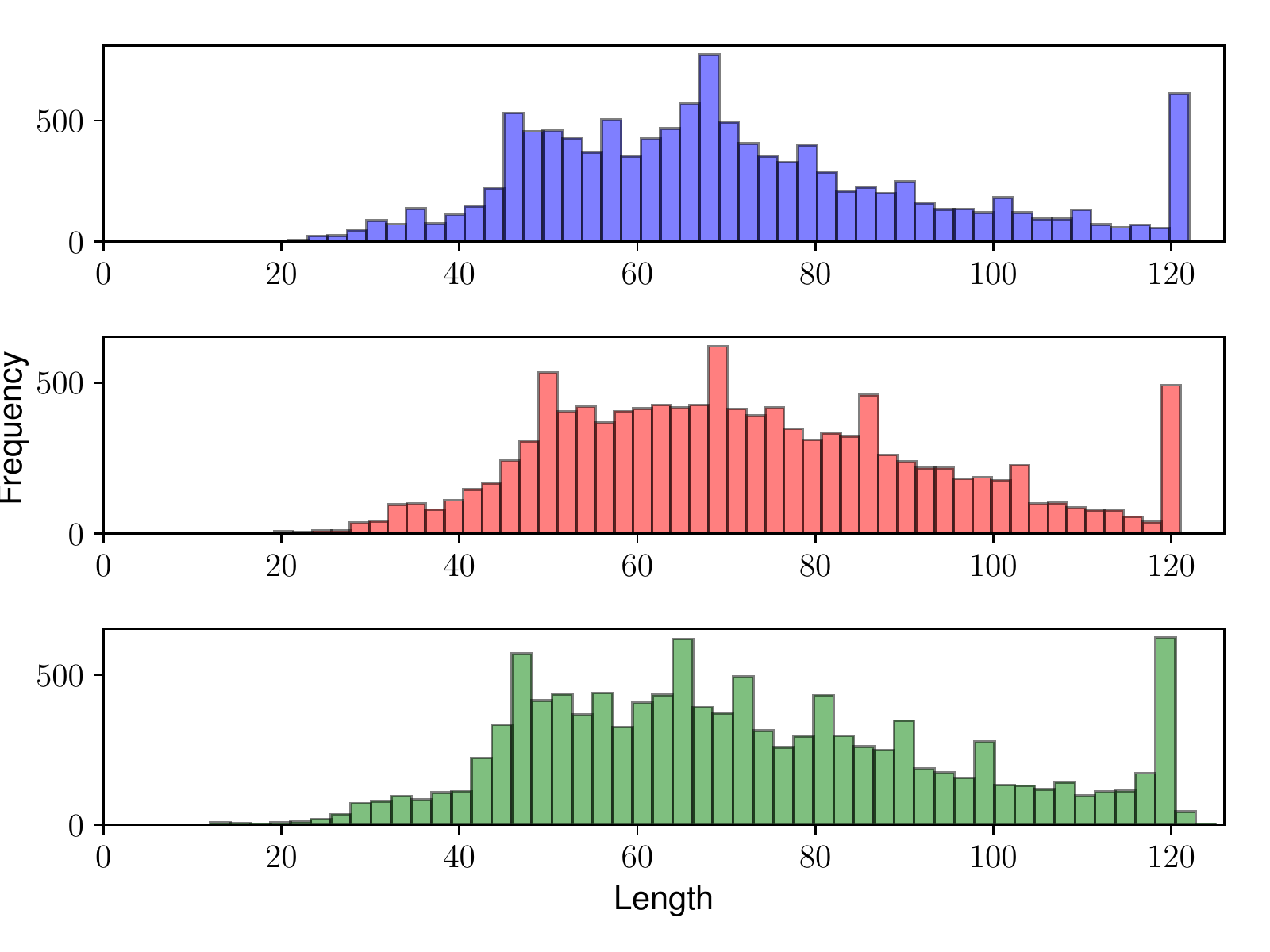} %
\caption{\ourmethod is capable of producing output of variable lengths, matching target sequence lengths and BART outputs for CNN-DM generations. From top to bottom: (a) distribution of target lengths; (b) distribution of predicted length by BART; (c) distribution of predicted length by \ourmethod.} \label{fig:generation-length} %
\end{figure}  

\begin{figure}[t]
\centering
\includegraphics[scale=0.5]{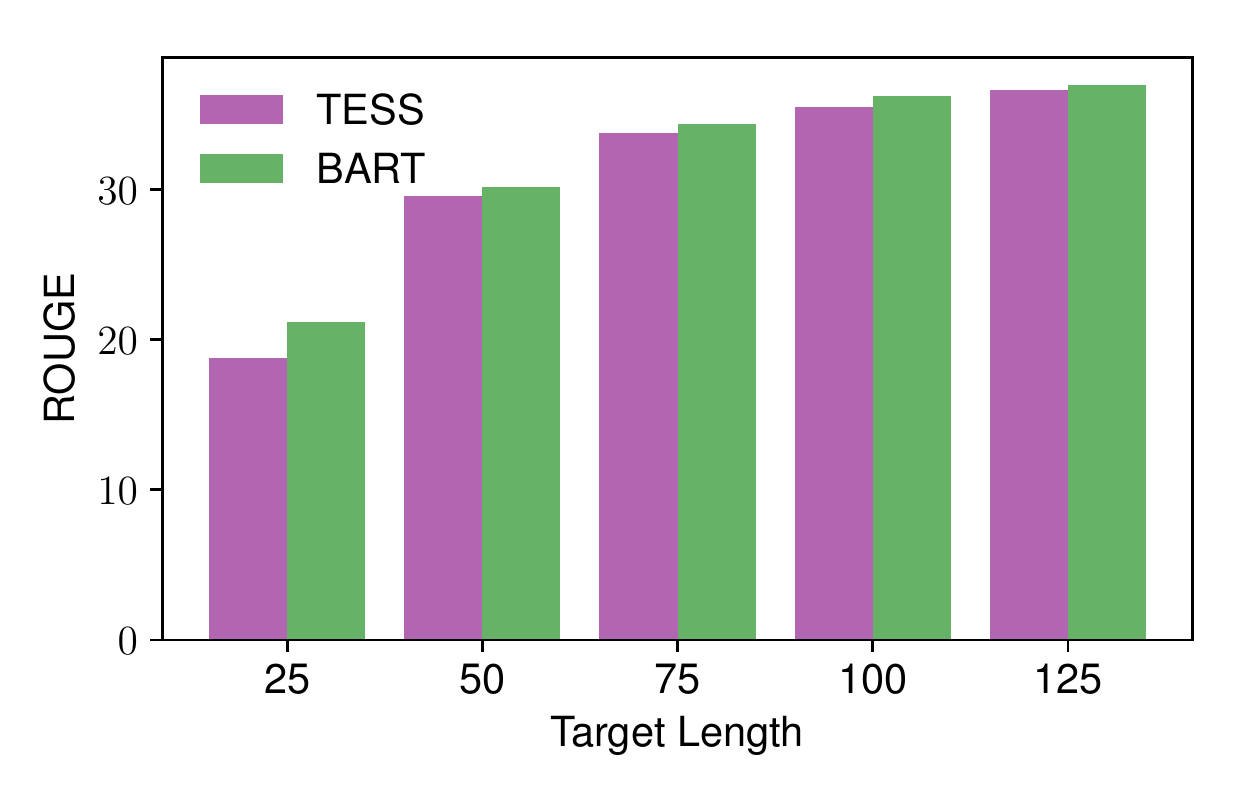}
\caption{Average ROUGE score (R1, R2, R-L) for \ourmethod vs BART for CNN-DM generations. Our method performs comparably to BART for different target lengths.} 
\label{fig:rouge-buckets}
\end{figure} 

\subsection{Variable Length Output}
\label{subsec:analysis-length}
Figure \ref{fig:generation-length} shows \ourmethod 
is capable of producing outputs of variable lengths that match the underlying distribution of sequence lengths in the gold data as well as BART outputs. 
We also evaluate generation quality for differing output lengths in Figure~\ref{fig:rouge-buckets}. We observe that \ourmethod is consistent with the BART baseline for variable target lengths, with longer generations matching BART's performance.


\subsection{Self-Conditioning}
\label{sec:ablations}
To examine the impact of self-conditioning, we compare our proposed method with the original strategy~\cite{chen2022analog} in text simplification and paraphrase generation tasks. As shown in Table~\ref{tab:self}, adding self-conditioning consistently improves results, with our variant delivering the best overall performance.

\subsection{Sampling Steps}

We also investigate the quality of \ourmethod generations on the suite of NLG tasks as well as MRPC by varying the number of sampling steps during inference.
As shown in Table \ref{tab:sampling-steps}, \ourmethod performs well with relatively few steps, with only a marginal drop in performance even when sampling steps are decreased from 1000 to 100. For classification tasks that involve shorter generation (MRPC), 10 sampling steps result in lossless quality. We also find that decreasing the number of sampling steps is possible in generative tasks like question generation. We provide results in Appendix~\ref{app:forwards}. Notably, it appears that the number of steps required correlates with the difficulty of the task: while classification tasks such as MRPC only require few steps, longer generation tasks such as CNN-DM require closer to 100 steps to achieve good performance.

\begin{table*}[t!]
\centering
\footnotesize
\begin{tabular}{lccccccc}
\toprule
\multirow{2}{*}{\textbf{Steps}} & {\textbf{QQP}} & {\textbf{NEWSELA-AUTO}} & {\textbf{QG}} & {\textbf{CNN-DM}} & \multicolumn{2}{c}{{\textbf{MRPC}}} \\
\cmidrule(lr){2-5} \cmidrule(lr){6-7} 
\multicolumn{5}{c}{\textbf{R-L}} & \textbf{Accuracy} & \textbf{F1} \\
\midrule
10 & \textbf{62.4} & 58.4 & 38.8 & 35.6 & \textbf{89.7} & \textbf{92.8} \\
100 & 62.0 & 59.1 & \textbf{38.9} & 39.6 & \textbf{89.7} & \textbf{92.8} \\
1000 & 62.2 & \textbf{59.4} & \textbf{38.9} & \textbf{39.7} & \textbf{89.7} & \textbf{92.8} \\
\bottomrule
\end{tabular}
\caption{Impact of number of sampling steps on performance. Our method achieves competitive results with as few as 10 or 100 steps on NLG tasks and 10 for MRPC.}
\label{tab:sampling-steps}
\end{table*}

\begin{figure}
    \centering
    \adjustbox{width=\linewidth}{
        \includegraphics{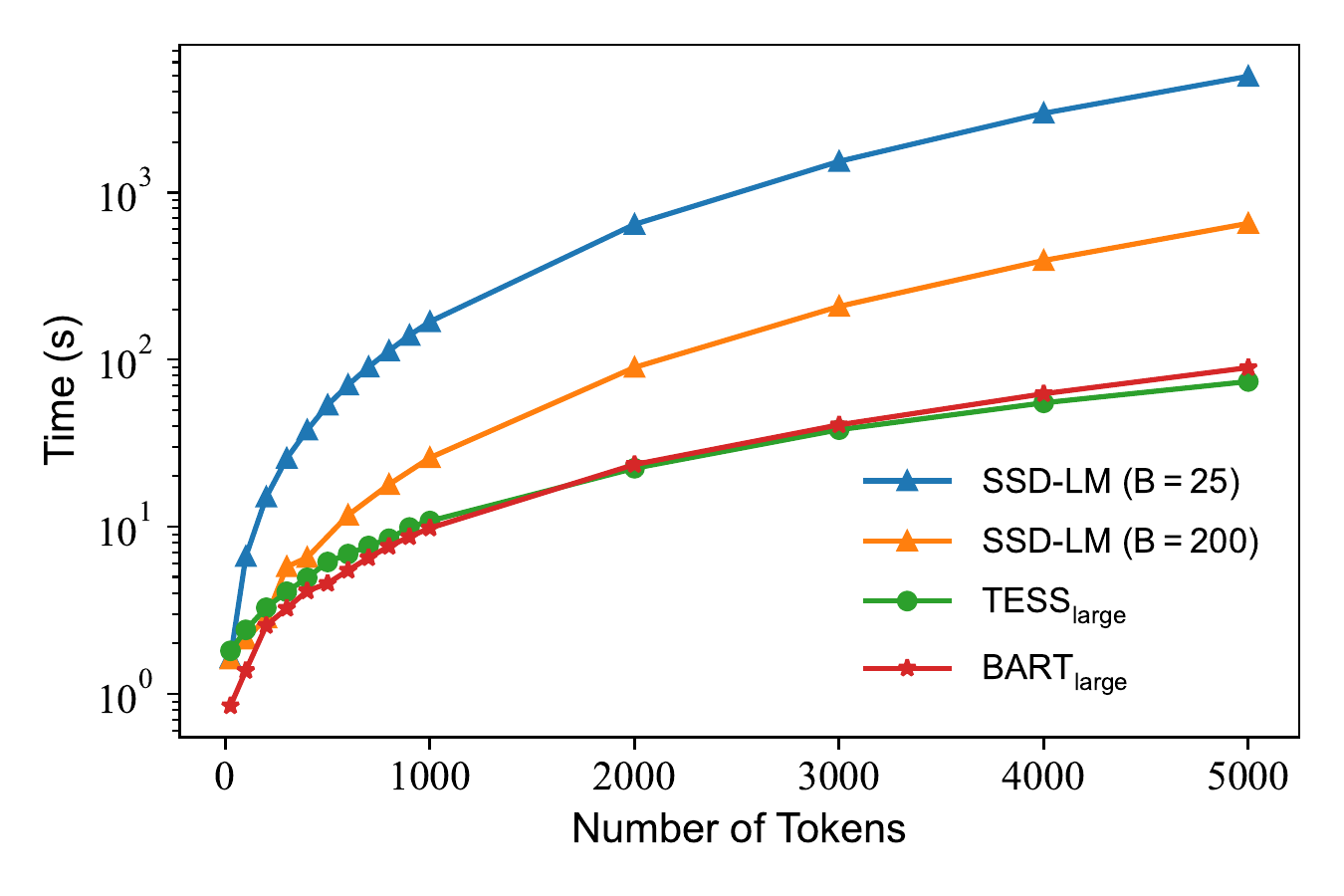}
    }
    \caption{Time taken to generate variable number of tokens with 100 diffusion steps. We report the average time over five runs. Diffusion-based models use RoBERTa\textsubscript{large} as their backbone. \ourmethod is substantially faster than SSD-LM. Notably, for 2000 tokens, it is even marginally faster than an equivalently-sized BART.}
    \label{fig:inference_runtime}
\end{figure}

\subsection{Sampling Speed}


We compare \ourmethod generation speed with other models in Figure~\ref{fig:inference_runtime}. We time how long decoding 25 to 5000 tokens takes given a context of 50 tokens and 100 diffusion steps. We find that \ourmethod is substantially faster than SSD-LM, especially when SSD-LM has to generate multiple blocks due to its limited block size. Notably, we find \ourmethod is faster, albeit marginally, than an equivalently-sized BART when generating more than 2000 tokens. We provide further details in Appendix~\ref{app:speed_exps}.

\subsection{\ourmethod vs other Diffusion Methods}

As shown in Section~\ref{sec:results}, \ourmethod outperforms other diffusion methods across several benchmarks. We believe this is due to a number of factors: (1) the simplex-based formulation being a more natural fit for language than embedding-based ones, allowing us avoid methods like clamping or an extra decoder for exiting the embedding space; (2) the simplex-based self-conditioning formulation, which we empirically show outperforms more standard self-conditioning methods (Table~\ref{tab:self}); (3) the use of a pretrained model - while \ourmethod can still outperform other methods without using a pretrained model (Table~\ref{tbl:question-generation}), being able to make use of pretrained models with relatively little extra training. 

\section{Related Work}
\label{sec:related}

\paragraph{Diffusion for continuous domains} 
Diffusion models were first proposed by ~\citet{sohl2015deep} 
and popularized by Denoising Diffusion Probabilistic Models (DDPMs) \citep{ho2020denoising}, which proposed a new parameterization that revealed an equivalence between ground-truth prediction and noise estimation. \citet{song2021scorebased} proposed an alternative stochastic differential equation interpretation of diffusion that involves the Stein score function. 
\citet{nichol2021improved} proposed a number of modifications to DDPMs,
which improved log-likelihood and reduced sampling steps. \citet{ho2022classifier} proposed classifier-free guidance, which allows for highly controllable generation without the need for an external classifier to guide the model score estimates.

\paragraph{Continuous diffusion for discrete domains} Following the success of diffusion models on continuous domains, there have been several attempts to apply diffusion on discrete data. \citet{li2022diffusionlm} applied diffusion on the latent token embedding space. Their resulting language model relies on word-level tokenization and works mostly on small datasets with a short sequence length of 64 tokens.

\citet{strudel2022self} used frozen pretrained word embedding with careful scaling to address the instability resulting from the competition between diffusion and reconstruction loss. However, their method does not allow the joint training of word embeddings, and the model was not evaluated on downstream NLP tasks. More recently,~\citet{dieleman2022continuous} attempted to learn the embedding and diffusion model jointly, still by performing diffusion in the embedding space. Other recent works have also applied diffusion on word embeddings to tackle sequence-to-sequence problems~\cite{gong2022diffuseq,yuan2022seqdiffuseq}. Concurrent to our work, \citet{ye2023dinoiser} proposed methods for manipulating the noise in the diffusion process during training and inference, yielding improved conditional text generation. Another concurrent work explores variational diffusion models for language modeling in embedding space~\citep{gulrajani2023likelihood}. However, they compare their models to $8\times$ smaller autoregressive models; our method obtains competitive performance with same-size autoregressive models.

Most relevant to our work,~\citet{han2022ssd} proposed a semi-autoregressive diffusion model which generates small blocks of 25 tokens from left to right, feeding them as additional context to generate next blocks. We extend their approach to fully non-autoregressive generation which substantially speeds up the inference time and incorporate an efficient self-conditioning method that exploits the semantics of the simplex space. \citet{han2023ssd2} similarly extends this approach, but focuses on showing the viability of simplex-based diffusion with large (> 1B parameter) models.

\paragraph{Discrete diffusion for discrete domains}
Unlike continuous diffusion models, discrete diffusion models maintain the discrete structure of the data domain and perform state transitions based on a probability matrix. Diffusion models with discrete state space were first explored by~\citet{sohl2015deep}, who proposed a framework for diffusion over binary random variables. Later, \citet{hoogeboom2021argmax} and \citet{austin2021structured} proposed discrete diffusion models for categorical random variables. However, these methods generally lag behind autoregressive models. More recently, \citet{reid2022diffuser} proposed Diffuser, which formulates a discrete diffusion process by modeling generation as a series of discrete edit operations. \ourmethod substantially outperforms Diffuser.

\section{Conclusion}
\label{sec:future}
We present \ourmethod, a new sequence-to-sequence diffusion model for language generation tasks that is fully non-autoregressive, works for long sequences compared to prior work, performs the diffusion process on the vocabulary logit space, and employs a new and efficient form of self-conditioning. \ourmethod outperforms strong autoregressive baselines as well as recent state-of-the-art text diffusion models on a wide variety of conditional language generation and language understanding tasks, while also being far more efficient than prior diffusion-based models.
Future work relies on pretraining our method combined with denoising and infilling objectives, which we hypothesize can provide further performance boosts to our text-to-text diffusion model.

\section*{Limitations}

\paragraph{Sampling speed} As seen in Figure~\ref{fig:inference_runtime}, \ourmethod is still slower than BART when generating $<$ 1000 tokens. We experimented with reducing the number of diffusion steps (see Table~\ref{tab:sampling-steps}), which can further speedup the generation. While in majority of tasks using just 10 steps provides promising results, it is not enough to achieve strong performance on more complex tasks such as summarization. Incorporating recent work in computer vision to accelerate sampling in diffusion-based models~\cite{song2023consistency} could result in further speedups in generation. 


\paragraph{Long sequences} Inference speed tests with SSD-LM and BART show that \ourmethod quickly dominates semi-autoregressive generation and outperforms BART at 2000 tokens. This result suggests that diffusion models have the potential of being faster than popular autoregressive models at long sequence lengths. In this work, we primarily used RoBERTa\textsubscript{base} models to facilitate fair comparison with existing baselines, which inevitably limited the size of the context window due to the absolute position embedding strategy employed by RoBERTa. We suspect that unlocking the full potential of diffusion-based language models may lie in the long sequence regime, which could involve scaling up the current models.

\section*{Ethics Statement}

Language models are known to produce toxic and biased content \citep{lmharms, sheng-etal-2021-societal}. While we explore an alternate modelling framework to that commonly used in prior studies on the toxicity of language models, there is little reason to suggest our models would not also contain these issues. However, given the greater controllability of the diffusion framework~\citep{li2022diffusion}, we hope future work explores how to make use of this controllability to reduce potential harms. Further examining how well results around toxic and harmful generations of autoregressive setups transfer to our setting may also aid in identifying future areas for improvement.

\section*{Acknowledgements}
We are grateful to Aman Madaan, Robin Strudel,
Sander Dieleman, Chris Dyer, Xiaochuang Han,
Sachin Kumar, Clara Meister, Sean Welleck, and
Andre Wibisono for helpful comments and discussions, and Sam Skjonsberg and the ReViz team at
AI2 for their support in managing experiments.

\bibliography{anthology,custom}

\clearpage
\appendix
\label{sec:appendix}

\section{Experiment Details} \label{app:experimental_detials}

\subsection{Dataset} 

NEWSELA-AUTO dataset~\citep{jiang2020neural} is based on a \citet{xu-etal-2015-problems} with revised alignment and improved quantity and quality. For question generation, we use the QUASAR-T dataset~\citep{dhingra2017quasar}; for summarization we use CNN-DailyMail~\citep{hermann2015teaching}. 

The GLUE benchmark~\citep{wang2018glue} is released under the Creative Commons License (CC BY 4.0). This benchmark consists of multiple datasets: SST-2~\citep{socher-etal-2013-recursive}, MNLI~\citep{williams-etal-2018-broad}, CoLA~\citep{warstadt-etal-2019-neural}, MRPC~\citep{dolan-brockett-2005-automatically}, QQP\footnote{\url{https://quoradata.quora.com/First-Quora-Dataset-Release-Question-Pairs}}, QNLI~\citep{rajpurkar-etal-2016-squad}, STS-B~\citep{cer-etal-2017-semeval}, and RTE, which is a combination of data from RTE1~\citep{dagan2005pascal}, RTE2~\citep{rte2}, RTE3~\citep{giampiccolo-etal-2007-third}, RTE5~\citep{Bentivogli09thefifth}. We download all datasets from the Hugging Face Datasets library \citep{lhoest-etal-2021-datasets}. Table~\ref{tab:seq-lengths} shows the sequence lengths for the source and target in each dataset, in number of tokens.

\begin{table}[ht] 
\centering
\footnotesize
\begin{tabular}{lcc}
\toprule
\textbf{Dataset} &  \textbf{Source} & \textbf{Target} \\
\midrule 
GLUE    &  128 & 5 \\
NEWSELA-AUTO & 128 & 128 \\
QQP & 100 & 85 \\
QG & 155 & 65 \\
CNN-DM & 392 & 120 \\
\bottomrule 
\end{tabular}
\caption{Sequence length of each dataset.}
\label{tab:seq-lengths}
\end{table}

\subsection{Baseline Details}

When prior published results are on the same dataset and metric, or when the codebase is not publicly available, we use available reported results to reduce compute costs. To provide a fair comparison when tuning baselines ourselves (SSD-LM and BART), we use the same tuning budget and compared them in the same setting (see Appendix~\ref{app:training_hyperparam}). Lastly, we report values using the same decoding strategy, i.e., the default setting \textit{without} minimum risk Bayes decoding (MBR).

For GENIE, we use the code and model weights provided by the authors\footnote{\url{https://github.com/microsoft/ProphetNet/tree/master/GENIE}} to evaluate GENIE using matched decoding settings to TESS.

\subsection{Training Hyperparameters}
\label{app:training_hyperparam}
Following~\citet{han2022ssd}, we initialized the model from a pretrained model and similarly found that it improved performance (See Table~\ref{tbl:question-generation}). We implemented our work using HuggingFace Transformers~\citep{wolf-etal2020transformers} and used the Huggingface Diffusers\footnote{\url{https://huggingface.co/docs/diffusers}} to build our diffusion pipeline. Our experiments are performed on 8 NVIDIA A6000/A100 GPUs. 

We trained our models and baselines with a learning rate of $3\mathrm{e}{-5}$ with the AdamW optimizer, with default parameters $\beta_1 = 0.9$, $\beta_2 = 0.999$, $\epsilon = 1\mathrm{e}{-8}$. We use a linear learning rate scheduler. We use the base model sizes for all experiments~\citep{wolf-etal2020transformers}. 


For SSD-LM, we use a block size of 25 following the original paper and the same number of diffusion steps during training and inference as our own models. We reuse the codebase provided by the authors and adapt it to support downstream tasks. We do not test SSD-LM on CNN-DM due to the difficulty of training SSD-LM on outputs involving multiple decoding blocks, requiring custom data preprocessing, and further algorithm tweaks to handle the long outputs.

For all generation tasks, we train our method and baselines for paraphrase generation, summarization, question generation, and text simplification for 90K, 120K, 120K, and 80K steps, respectively.
We set the number of warmup steps to 2000 for all generation tasks.
For the experiments on GLUE, we set the number of warm-up steps to 500. 
We trained the models on larger datasets in GLUE for 25K steps; for smaller datasets, we use 12K steps. We then evaluate the models every 1K steps and report the results on the checkpoint obtaining the best results on the development set. 
We found that the training time of each model is roughly similar: with equivalent configurations on a single GPU on the QQP dataset, TESS achieves 1.7 train steps per second; SSD-LM, 1.8; BART, 1.4, using PyTorch 2.0.

\subsection{Evaluation Package Details}

We use the following packages for calculating the given metric:
\begin{itemize}
    \item \textbf{BLEU}: We use the \texttt{sacrebleu} package \citep{post-2018-call}, v2.3.1.
    \item \textbf{ROUGE}: We use the \texttt{rouge-score} package, v0.2.1.\footnote{\url{https://github.com/google-research/google-research/tree/master/rouge}}
    \item \textbf{Mauve}: We use the \texttt{mauve-text} package \citep{pillutla2021mauve}, v0.3.0.
    \item \textbf{BERTScore}: We use the \texttt{bert-score} package \citep{zhang2019bertscore}, v0.3.12.
\end{itemize}

For other metrics, we use our own implementations (usually heavily based on a reference implementation), which will be open-sourced.

\subsection{Inference Speed Experiments}
\label{app:speed_exps}
For the inference speed numbers reported in Table~\ref{fig:inference_runtime}, we run all experiments on a single 80 GB NVIDIA A100 GPU. We use an adapted version of the SSD-LM inference script provided by the authors in their public repository, removing logging and initializing tensors on-device to avoid expensive \texttt{.to()} calls. For BART\textsubscript{large}, we use the transformers library \citep{wolf-etal2020transformers}. We alter all models to allow sequence lengths over 512 tokens by resizing the position embeddings matrix. We use the following context: \textit{``A man of innumerable personalities and powers vs. the most powerful artificial intelligence in this universe: Legion vs. Nimrod! With Nightcrawler in Orchis clutches, David Haller and his allies will have to confront the mastermind who".} We report the exact timings and standard deviations in Table~\ref{tab:seq-lengths2}.

\section{Sampling Steps} \label{app:forwards}

We performed additional ablations on the relationship between the number of sampling steps and performance on question generation. 

\begin{table}[ht] 
\centering
\footnotesize
\begin{tabular}{lllc}
\toprule
\textbf{Model} & \textbf{Steps} & \textbf{Forwards} & \textbf{R-L} \\
\midrule 
BART & --- & 74 & 38.8 \\
SSD-LM & 10 & 50 & 33.0 \\
SSD-LM & 100 & 500 & 36.7 \\
SSD-LM & 1000 & 5000 & 38.5 \\
TESS & 10 & 10 & 38.8 \\
TESS & 100 & 100 & \bf{38.9} \\
TESS & 1000 & 1000 & \bf{38.9} \\
\bottomrule 
\end{tabular}
\caption{Sampling step ablation on question generation.}
\label{tab:forwards}
\end{table}

We note forward passes are not directly comparable between AR and NAR models: AR models may use more or less forwards than NAR models depending on the number of tokens generated, and each forward pass of the AR model involves a differing number of tokens. Here, we use 74, as this is the average number of BART tokens in question generation responses.

Overall, we observe that SSD-LM's performance drops significantly with fewer diffusion steps, while TESS remains largely unaffected. Notably, TESS achieves parity with BART with only 10 sampling steps.

\section{Inference Step} \label{app:inference-step}
In the typical variant of DDPM, the model predicts the added noise  $\bm{\epsilon_{\theta}}(\mathbf{x}_t, t)$ instead of original signal and the DDPM inference step~\citep{ho2020denoising} is as follows:\footnote{Following~\citet{han2022ssd} we drop the additional noise term $\sigma_t \bm{z}$, where $\bm{z} \in \mathcal{N}(0, \bm{I})$.}
\begin{align}
\mathbf{x}_{t-1} = \frac{1}{\sqrt{\alpha_t}}\left( \mathbf{x}_{t} - \frac{1-\alpha_t}{\sqrt{1-\bar{\alpha}_t}} \bm{\epsilon_\theta}(\mathbf{x}_{t}, t)\right)
\label{eq:ddpm-inference}
\end{align}
Since we work with the variant predicting the signal itself, we substitute \eqref{eq:ddpm-forward} into \eqref{eq:ddpm-inference}, obtaining:
\begin{align}
\mathbf{x}_{t-1} = \frac{1}{\sqrt{\alpha_t}}\left(\sqrt{\bar{\alpha}_t} \mathbf{x}_{0}  - \frac{\alpha_t-\bar{\alpha}_t}{\sqrt{1-\bar{\alpha}_t}} \bm{\epsilon_\theta}(\mathbf{x}_{t}, t)\right).
\end{align}
Given that $\bar{\alpha_t} = \bar{\alpha}_{t-1} \alpha_t$, we arrive at: 
\begin{align}
\mathbf{x}_{t-1} = \sqrt{\bar{\alpha}_{t-1}} \mathbf{x}_{0}  - \frac{\sqrt{\alpha_t-\bar{\alpha}_t}\sqrt{\alpha_t-\bar{\alpha}_t}}{\sqrt{\alpha_t}\sqrt{1-\bar{\alpha}_t}} \bm{\epsilon_\theta}(\mathbf{x}_{t}, t).
\label{eq:our-inference2}
\end{align}

Using a cosine schedule for $\bar{\alpha}_t$~\citep{nichol2021improved}, $\sqrt{{(\alpha_t-\bar{\alpha}_t)}/{(1-\bar{\alpha}_t)}} \geq 0.98$ for 98\% $t \in (1,T)$, with some outliers as $t \rightarrow 0$ and $t\rightarrow T$~\citep{han2022ssd}. Thus, with the approximation $\sqrt{{(\alpha_t-\bar{\alpha}_t)}/{(1-\bar{\alpha}_t)}} \approx 1$, Equation~\eqref{eq:our-inference2} further simplifies into: 
\begin{align}
\mathbf{x}_{t-1} = \sqrt{\bar{\alpha}_{t-1}} \mathbf{x}_{0}  - \sqrt{1-\bar{\alpha}_{t-1}} \bm{\epsilon_\theta}(\mathbf{x}_{t}, t).
\end{align}

In our case, the signal $\mathbf{x}_{t}$ is the simplex $\mathbf{S}_t$, with $\hat{\mathbf{S}}_{\bm{\theta}}$ as the model prediction of the ground-truth. Adjusting the above with this notation, we recover Equation~\eqref{eq:rev_diff}:
\begin{align*}
\mathbf{S}_{t - 1} = \sqrt{\bar{\alpha}_{t-1}} \hat{\mathbf{S}}_{\bm{\theta}}(\mathbf{S}_t, t) + \sqrt{1-\bar{\alpha}_{t-1}} \bm{\epsilon}_t.
\end{align*}

\section{Qualitative Examples}

We show randomly chosen example outputs from \ourmethod model and BART, the strongest baseline, on the summarization task in Table~\ref{tab:qualitative}. Qualitatively, we observe that \ourmethod is capable of generating natural samples that are often indistinguishable from those of BART.

\begin{table*}[ht] 
\centering
\footnotesize
\setlength\tabcolsep{2pt}
\adjustbox{max width=\textwidth}{
\begin{tabular}{lcccccccccccccccc}
\toprule
 & & \multicolumn{15}{c}{\textbf{Number of Tokens}} \\
\cmidrule{3-17} \textbf{Model} & \textbf{Blocks} & \textbf{25} & \textbf{100} & \textbf{200} & \textbf{300} & \textbf{400} & \textbf{500} & \textbf{600} & \textbf{700} & \textbf{800} & \textbf{900} & \textbf{1000} & \textbf{2000} & \textbf{3000} & \textbf{4000} & \textbf{5000}  \\
\midrule 
\textbf{SSD-LM} & 25 & 1.6$_{0.3}$ & $6.6_{0.3}$ & 15.0$_{0.3}$ & 25.6$_{0.3}$ & 37.8$_{0.3}$ & 53.4$_{0.4}$ & 70.4$_{0.0}$ & 90.1$_{0.0}$ & 112.7$_{0.0}$ & 139.4$_{0.0}$ & 168.4$_{0.0}$ & 643.2$_{0.2}$ & 1531.5$_{0.6}$ & 2933.4$_{0.9}$ & 4945.1$_{2.7}$ \\
\textbf{SSD-LM} & 200 & 1.6$_{0.3}$ & 2.1$_{0.3}$ & 2.8$_{0.3}$ & 5.8$_{0.3}$ & 6.5$_{0.3}$ & - & 11.7$_{0.3}$ &- & 17.9$_{0.0}$ & - & 25.8$_{0.0}$ & 89.6$_{0.0}$ & 207.8$_{0.0}$ & 391.1$_{0.2}$ & 652.2$_{0.2}$ \\
\textbf{BART\textsubscript{large}} & - & 0.8$_{0.2}$ & 1.4$_{0.2}$ & 2.6$_{0.2}$ & 3.2$_{0.2}$ & 4.1$_{0.2}$ & 4.6$_{0.2}$ & 5.5$_{0.0}$ & 6.5$_{0.0}$ & 7.6$_{0.0}$ & 8.7$_{0.0}$ & 9.8$_{0.0}$ & 23.5$_{0.2}$ & 40.6$_{0.1}$ & 62.4$_{0.1}$ & 89.4$_{0.4}$ \\
\textbf{TESS} & - & 1.8$_{0.3}$ & 2.4$_{0.3}$ & 3.3$_{0.3}$ & 4.1$_{0.3}$ & 4.9$_{0.3}$ & 6.1$_{0.3}$ & 6.9$_{0.0}$ & 7.7$_{0.0}$ & 8.5$_{0.0}$ & 9.9$_{0.0}$ & 10.8$_{0.0}$ & 22.3$_{0.0}$ & 38.0$_{0.0}$ & 55.0$_{0.0}$ & 73.8$_{0.0}$ \\
\bottomrule 
\end{tabular}}
\caption{Time taken to generate the given number of tokens with a 50-token prefix in seconds. All models use 100 diffusion steps and RoBERTa\textsubscript{large} as the underlying model. All values are the average over 5 runs, with standard deviations given as subscripts; standard deviations less than $0.05$ appear as $0.0$.}
\label{tab:seq-lengths2}
\end{table*}

    \begin{table*}
\tiny
\centering
\setlength\tabcolsep{3pt}
\setlength{\extrarowheight}{0pt}
\addtolength{\extrarowheight}{\aboverulesep}
\addtolength{\extrarowheight}{\belowrulesep}
\setlength{\aboverulesep}{0pt}
\setlength{\belowrulesep}{0pt}
\arrayrulecolor{black}
\begin{adjustbox}{max width=\linewidth}
\begin{tabular}{@{}ccc@{}}
\toprule
\textbf{Gold}                                                                                                                                                                                                                                                                                                                                                      & \textbf{BART}                                                                                                                                                                                                                                                                                                                                               & \textbf{Ours}                                                                                                                                                                                                                                                                                                                                                                                                         \\ 
\hline
\begin{tabular}[c]{@{}p{0.33\textwidth}@{}}Membership gives the ICC jurisdiction over alleged crimes committed in Palestinian territories since last June.\\Israel and the United States opposed the move, which could open the door to war crimes investigations against Israelis.\end{tabular}                                                                                                                    & \begin{tabular}[c]{@{}p{0.33\textwidth}@{}}The Palestinian Authority officially becomes the 123rd member of the International Criminal Court.\\It gives the court jurisdiction over alleged crimes committed in Palestinian territories.\\Israel and the United States, neither of which is an ICC member, opposed the Palestinians' efforts to join the body.\end{tabular} & \begin{tabular}[c]{@{}p{0.33\textwidth}@{}}The Palestinian Authority is the 123rd member of the International Criminal Court.\\The move gives it jurisdiction over alleged crimes in Palestinian territories.\\Israel and the United States oppose the Palestinians' efforts to join it.\end{tabular}                                                                                                                                 \\ 
\hline
\begin{tabular}[c]{@{}p{0.33\textwidth}@{}}Theia, a bully breed mix, was apparently hit by a car, whacked with a hammer and buried in a field.\\"She's a true miracle dog and she deserves a good life," says Sara Mellado, who is looking for a home for Theia.\end{tabular}                                                                                                                                       & \begin{tabular}[c]{@{}p{0.33\textwidth}@{}}A stray pooch in Washington State has used up at least three of her own injuries.\\Theia, a bully breed mix, was hit by a car and then buried in a field.\\She has been receiving care at the Veterinary Teaching Hospital.\end{tabular}                                                                                         & \begin{tabular}[c]{@{}p{0.33\textwidth}@{}}A stray dog was found on a farm in Washington State.\\The dog has used up at least three of her own after being hit in the head by a car and being buried in a field.\\Theia has a dislocated jaw, leg injuries and needs surgery to help breathe.\end{tabular}                                                                                                                            \\ 
\hline
\begin{tabular}[c]{@{}p{0.33\textwidth}@{}}Mohammad Javad Zarif has spent more time with John Kerry than any other foreign minister.\\He once participated in a takeover of the Iranian Consulate in San Francisco.\\The Iranian foreign minister tweets in English.\end{tabular}                                                                                                                                   & \begin{tabular}[c]{@{}p{0.33\textwidth}@{}}Mohammad Javad Zarif is the Iranian foreign minister.\\He has been John Kerry's opposite number in securing a breakthrough in nuclear talks.\end{tabular}                                                                                                                                                                        & \begin{tabular}[c]{@{}p{0.33\textwidth}@{}}Mohammad Javad Zarif is now Iran's foreign minister.\\He has been John Kerry's opposite number in securing a breakthrough in nuclear talks.\end{tabular}                                                                                                                                                                                                                                   \\ 
\hline
\begin{tabular}[c]{@{}p{0.33\textwidth}@{}}17 Americans were exposed to the Ebola virus while in Sierra Leone in March.\\Another person was diagnosed with the disease and taken to hospital in Maryland.\\National Institutes of Health says the patient is in fair condition after weeks of treatment.\end{tabular}                                                                                               & \begin{tabular}[c]{@{}p{0.33\textwidth}@{}}One of the five had a heart-related issue on Saturday and has been discharged.\\The others have already gone home.\end{tabular}                                                                                                                                                                                                  & \begin{tabular}[c]{@{}p{0.33\textwidth}@{}}Five Americans were monitored for three weeks at an Omaha hospital.\\One of the five had a heart-related issue on Saturday.\\The others have already gone home.\\They were exposed to Ebola in Sierra Leone, but none developed the deadly virus.\end{tabular}                                                                                                                             \\ 
\hline
\begin{tabular}[c]{@{}p{0.33\textwidth}@{}}Student is no longer on Duke University campus and will face disciplinary review.\\School officials identified student during investigation and the person admitted to hanging the noose, Duke says.\\The noose, made of rope, was discovered on campus about 2 a.m.\end{tabular}                                                                                        & \begin{tabular}[c]{@{}p{0.33\textwidth}@{}}A student admitted to hanging a noose made of rope from a tree near a student union.\\Duke didn't identify the student, citing federal privacy laws.\\The incident is one of several recent racist events to affect college students.\end{tabular}                                                                               & \begin{tabular}[c]{@{}p{0.33\textwidth}@{}}A Duke student admitted to hanging a noose from a tree.\\The private school didn't identify the student, citing federal privacy laws.\\The student is no longer on campus and will face student conduct review.\end{tabular}                                                                                                                                                               \\ 
\hline
\begin{tabular}[c]{@{}p{0.33\textwidth}@{}}College-bound basketball star asks girl with Down syndrome to high school prom.\\Pictures of the two during the "prom-posal" have gone viral.\end{tabular}                                                                                                                                                                                                               & \begin{tabular}[c]{@{}p{0.33\textwidth}@{}}Eastern High School basketball player Trey Moses asked his girlfriend to be his prom date.\\Ellie Meredith, a freshman with Down syndrome, has struggled with friendships since elementary school.\\A special program at Eastern has made things easier for her, her mom says.\end{tabular}                                      & \begin{tabular}[c]{@{}p{0.33\textwidth}@{}}College basketball player and high school freshman picked Ellie Meredith as his prom date.\\The photos have gone viral on social media.\end{tabular}                                                                                                                                                                                                                                       \\ 
\hline
\begin{tabular}[c]{@{}p{0.33\textwidth}@{}}Amnesty's annual death penalty report catalogs encouraging signs, but setbacks in numbers of those sentenced to death.\\Organization claims that governments around the world are using the threat of terrorism to advance executions.\\The number of executions worldwide has gone down by almost 22\% compared with 2013, but death sentences up by 28\%.\end{tabular} & \begin{tabular}[c]{@{}p{0.33\textwidth}@{}}Amnesty International says governments are using the death penalty to advance executions.\\At least 607 people were executed around the world in 2014, compared to 778 in 2013.\end{tabular}                                                                                                                                     & \begin{tabular}[c]{@{}p{0.33\textwidth}@{}}Amnesty International released its annual report on the death penalty.\\At least 607 people were executed in 2014, compared to 778 in 2013.\\Report cites Pakistan lifting a six-year moratorium on the execution of civilians.\\China has used the death penalty as a tool in its "Strike Hard" campaign against terrorism in the restive far-western province of Xinjiang.\end{tabular}  \\ 
\hline
\begin{tabular}[c]{@{}p{0.33\textwidth}@{}}Andrew Getty's death appears to be from natural causes, police say, citing coroner's early assessment.\\In a petition for a restraining order, Getty had written he had a serious medical condition.\\Police say this is not a criminal matter at this time.\end{tabular}                                                                                                & \begin{tabular}[c]{@{}p{0.33\textwidth}@{}}Andrew Getty appears to have died of natural causes, police say.\\The coroner's preliminary assessment is there was no foul play involved in his death.\\He was the grandson of oil tycoon J. Paul Getty.\end{tabular}                                                                                                           & \begin{tabular}[c]{@{}p{0.33\textwidth}@{}}Coroner's preliminary assessment says there was no foul play involved in the death of Getty.\\Andrew Getty, 47, had "several health issues," detective says.\\There is no criminal investigation underway, detective says.\end{tabular}                                                                                                                                                    \\ 
\hline
\begin{tabular}[c]{@{}p{0.33\textwidth}@{}}Once a super typhoon, Maysak is now a tropical storm with 70 mph winds.\\It could still cause flooding, landslides and other problems in the Philippines.\end{tabular}                                                                                                                                                                                                   & \begin{tabular}[c]{@{}p{0.33\textwidth}@{}}Maysak has lost a lot of steam as it spins west in the Pacific Ocean.\\It boasts steady winds of more than 70 mph (115 kph) and gusts up to 90 mph.\end{tabular}                                                                                                                                                                 & \begin{tabular}[c]{@{}p{0.33\textwidth}@{}}Maysak gained super typhoon status thanks to sustained 150 mph winds.\\Authorities have forbanned outdoor activities like swimming, surfing, diving and boating in some locales.\end{tabular}                                                                                                                                                                                              \\
\bottomrule
\end{tabular}
\end{adjustbox}
\caption{Randomly chosen samples generated on the CNN-DM dataset by BART and \ourmethod.}
\label{tab:qualitative}
\arrayrulecolor{black}
\end{table*}

\end{document}